\pdfoutput=1

\documentclass[11pt]{article}

\usepackage{EMNLP2023}

\usepackage{times}
\usepackage{latexsym}

\usepackage[T1]{fontenc}

\usepackage[utf8]{inputenc}

\usepackage{microtype}

\usepackage{inconsolata}

\usepackage{gb4e}
\noautomath
\usepackage{multirow}
\usepackage{multicol}
\usepackage{tikz}
\usepackage{adjustbox}
\usepackage{booktabs}
\usepackage{amsfonts}
\usepackage{cuted}
\usepackage{mathtools,amssymb}
\usepackage{forest}
\usepackage{makecell}
\usepackage{circuitikz}
\usepackage{xcolor}         
\usepackage{colortbl}

\title{Robust Generalization Strategies for Morpheme Glossing in an Endangered Language Documentation Context}

\author{Michael Ginn \and Alexis Palmer \\
  University of Colorado  \\
  \texttt{michael.ginn@colorado.edu} \and
  \texttt{alexis.palmer@colorado.edu} \\}

\begin{document}
\maketitle
\begin{abstract}
Generalization is of particular importance in resource-constrained settings, where the available training data may represent only a small fraction of the distribution of possible texts. We investigate the ability of morpheme labeling models to generalize by evaluating their performance on unseen genres of text, and we experiment with strategies for closing the gap between performance on in-distribution and out-of-distribution data. Specifically, we use weight decay optimization, output denoising, and iterative pseudo-labeling, and achieve a 2\% improvement on a test set containing texts from unseen genres. All experiments are performed using texts written in the Mayan language Uspanteko.
\end{abstract}

\section{Introduction}
With over half of the world's languages endangered \citep{seifart_language_2018}, language documentation is one of several strategies for preservation. Traditionally, many documentation projects have aimed to create grammatical descriptions, dictionaries, and annotated text corpora, in the form of interlinear glossed text (IGT; see section \ref{sec:igt}). The annotated texts can be used in the creation of reference tools and pedagogical materials, as well as providing input data for downstream tasks such as machine translation \citep{Zhou2019UsingIG}, morphological paradigm induction \citep{moeller-etal-2020-igt2p}, dependency parsing \citep{Georgi2012ImprovingDP}, and other tasks \citep{Georgi2016FromAT}, making it particularly valuable for low-resource languages.

Annotation of large corpora can be time-consuming and monotonous, so there is a desire for systems to automatically produce IGT, annotating plain text with labels describing the part-of-speech, morphology, and syntax of each word in the corpus \citep{ginn-etal-2023-findings}. These systems can be used in conjunction with human annotators to create annotated corpora rapidly, ensuring consistency and reducing the amount of human effort required. Importantly, reducing annotation time also frees up language experts to work on other types of language preservation or revitalization activities.

However, generalization for automated annotation systems remains a critical problem. Pre-existing corpora of annotated text are often small, contain transcriptions of spoken language from a small number of distinct speakers, and focus on specific types of language such as story-telling and oration.
Thus, systems trained on these corpora have difficulty generalizing to out-of-distribution (OOD) language, limiting their utility and robustness. 

As acquiring additional annotated data is generally expensive and difficult, it is preferable to design models that generalize well to OOD data. In this work, we design models for one type of text annotation: labeling each morpheme in a text with its grammatical function. We envision these models being used alongside human annotators to provide suggestions and annotate text more quickly and consistently than by human labeling alone.

We examine three strategies to improve the robustness of these morpheme labeling models with limited data:
\begin{enumerate}
    \item We optimize weight decay to improve generalization of large models.
    \item We apply a separate denoiser model to improve performance on out-of-vocabulary inputs.
    \item We apply self-supervised learning on unlabeled texts.
\end{enumerate}

Our experiments evaluate model performance on texts of different genres than the texts in the training set, in order to investigate their ability to generalize to future, out-of-distribution texts. We find that these strategies achieve small performance improvements on in- and out-of-distribution texts, and may be valuable for building more robust morpheme labeling models. Our code is available on GitHub.\footnote{\url{https://github.com/michaelpginn/igt-glossing}}

\section{Background}
\subsection{Interlinear Glossed Text}\label{sec:igt}
In language documentation projects, annotated text typically uses a standardized format such as Interlinear Glossed Text (IGT) \citep{comrie2008leipzig}, although the exact glossing conventions vary across projects. An example IGT sentence in Uspanteko is provided in \ref{igt}. 

\begin{small}
  \begin{exe}
    \ex 
    \gll Ti- j- ya' -tq -a' juntiir \\
    \textsc{Inc}- \textsc{E3s}- \textit{give} -\textsc{Pl} -\textsc{Enf} \textit{everything} \\
    \trans They give us everything \\
    \citep{pixabaj2007text}
    \label{igt}
  \end{exe}
  \end{small}

The first line of the example is a \textbf{transcription} in the target language. Words may be transcribed as-is, or divided into morphemes (meaning-bearing units of language), as in the example.

The second line of the example gives a \textbf{gloss} for each morpheme. Glosses typically indicate either the translation of a morpheme or its grammatical function. For example, the \textit{-tq-} morpheme is glossed as \textsc{Pl} (plural). Stem morphemes, such as \textit{ya'}, are glossed either with their translation (as here) or with a gloss indicating the stem type (such as \textsc{VT} for "transitive verb"). Our systems gloss stems using the latter approach.

The third line provides a translation of the sentence in a high-resource language, such as English.

Although there exist some large mixed-language corpora of IGT such as ODIN \citep{lewis_developing_2010} and IMTVault \citep{nordhoff2022imtvault}, the availability of IGT data is limited. For many languages, only small IGT corpora are available, and different corpora may (and do) use various annotation conventions. Depending on the wishes of the language community, such corpora may or may not be available for wider use or distribution.

\subsection{Task}
In this research, the task our systems address is to predict the gloss line of IGT given the transcription, segmented into morphemes. Each morpheme should be glossed with its grammatical function; to keep the output vocabulary small, we gloss stems with part-of-speech labels instead of translations. Using the example in \autoref{igt}, the input to the system would be the sequence $$\textnormal{"Ti j ya' tq a' [SEP] juntiir"}$$ and the intended output would be $$\textsc{"Inc E3s VT Pl Enf [SEP] Adv"}$$

where stems such as "ya'" and "juntiir" are glossed with the stem type, here \textsc{VT} for transitive verb and \textsc{Adv} for adverb.

\subsection{Related Work}

Existing scholarship has used a variety of approaches for automated gloss prediction, including rule-based methods \citep{bender2014learning}, active learning \citep{palmer_computational_2010, palmer2009evaluating}, conditional random fields \citep{moeller_automatic_2018, mcmillan-major_automating_2020}, and neural models \citep{moeller_automatic_2018, zhao_automatic_2020}. \citet{ginn2023taxonomic} experiment with morphologically-inspired loss functions to improve low-resource glossing models.
However, to our knowledge, there has been no evaluation or experimentation with generalization of these models. 

One of the 2023 SIGMORPHON shared tasks involved creating models for automated gloss prediction \citep{ginn-etal-2023-findings}, with participant systems employing strategies such as leveraging the translation line for stem glossing \citep{okabe-yvon-2023-lisn}, pretraining on large multilingual corpora \citep{he-etal-2023-sigmorefun}, and straight-through gradient estimation \citep{girrbach-2022-sigmorphon}.

Although the majority of machine learning research has traditionally evaluated models on in-distribution data, the ability to generalize to out-of-distribution data is desirable for natural language models \citep{linzen-2020-accelerate, lake_ullman_tenenbaum_gershman_2017}. This is particularly important for low-resource languages where collecting a wide distribution of data can be expensive or even infeasible.

\section{Data \& Methodology}
\subsection{Data}
This work uses a corpus of IGT data for Uspanteko, a low-resource Mayan language, originally from the OKMA documentation project \citep{pixabaj2007text} and adapted by \citet{palmer2009evaluating}. Morphemes are glossed with 68 different labels, plus a separator label. Each text was produced through recording speakers, transcribing text, and glossing with morpheme tags and translations. The corpus used includes 17 different speakers.

For this research, we experiment with generalization to unseen texts that represent different genres of text. This consideration is very practical for documentation projects, where the available training corpora are often the result of a single data collection project, and sometimes contain only one or two genres or registers of speech.

The Uspanteko corpus contains 27 texts in four different genres: stories, histories, personal anecdotes, and advice. We use the story and history texts as our in-distribution (ID) data, as we hypothesize that stories and histories have similar grammar and vocabulary. We use personal anecdotes and advice as our out-of-distribution (OOD) data. One intuitive difference between these sets is that stories and histories tend to talk about others, while an anecdote is about the speaker (and thus tends to use first-person voice) and advice is about the listener (second-person voice). There is only one instance where a document created by the same speaker appears in both the ID and OOD splits.

We randomly divide the ID data into training and evaluation sets and divide the OOD data into evaluation and final testing sets. The splits are listed in \autoref{tab:data_splits}.

\begin{table}[h]
    \centering
    \def\arraystretch{1.5}
    \begin{tabular}{|c|c c|}
    \hline
        Set & Genre(s)  & \# Sentences \\
        \hline
        Training & Story, History & 5049 \\
        Eval (ID) & Story, History & 2128  \\
        Eval (OOD) & Personal, Advice & 2128 \\
        Test (OOD) & Personal, Advice &  2128 \\
        \hline
    \end{tabular}
    \caption{Data splits, including in-distribution (ID) and out-of-distribution (OOD) data}
    \label{tab:data_splits}
\end{table}

To verify that these splits represent accurate distributions, we pretrained a masked language model on the training set (described in \autoref{Pretraining}) and calculated the perplexity for the ID and OOD eval sets.

\begin{table}[h]
    \centering
    \def\arraystretch{1.5}
    \begin{tabular}{|c|c|}
    \hline
        Set & Perplexity \\
        \hline
        Eval (ID) & 77.78 \\
        Eval (OOD) & 94.03 \\
        \hline
    \end{tabular}
    \caption{Perplexity of pretrained language model on data splits}
    \label{tab:pretraining_results}
\end{table}

Of course, genre and register only represent one form of out-of-distribution data. Data may also be out-of-distribution due to different speakers, dialects of a language, time period, and other factors.

All transcription data is segmented into morphemes. Thus, the task is to predict a gloss label for each morpheme in a sequence.

\subsection{Pretraining} \label{Pretraining}
Existing pretrained models are rarely available for low-resource languages such as Uspanteko. Thus, we pretrain a new masked language model (MLM) on the training set before fine-tuning to the task at hand (on the same data set). We use a smaller variation of the RoBERTa architecture \citep{liu_roberta_2019} to prevent over-fitting and reduce resources used. The model uses 3 hidden layers, hidden layers of size 100, and 5 attention heads, as in \citet{gessler_microbert_2023}, and we found in preliminary experiments that there is no significant difference in performance from a full-size RoBERTa model.

The model is pretrained using the parameters listed in \autoref{tab:traininghyper}. We employ a dynamic masking strategy \citep{liu_roberta_2019} where 15\% of tokens are masked, of which 80\% use a \textit{MASK} token, 10\% use a random token, and 10\% use the original token.

\begin{table}[!h]
    \centering
    \def\arraystretch{1.5}
    \begin{tabular}{l c}
        \toprule
        Parameter & Value \\
        \midrule
        Optimizer & AdamW  \\
        $\beta_1$ & 0.9 \\
        $\beta_2$ & 0.999 \\
        $\epsilon$ & $1\text{E}{-8}$ \\
        Weight decay & 0 \\
        Batch size & 64 \\
        Gradient accumulation steps & 3 \\
        Epochs & 50 \\
        GPU & NVIDIA V100 \\
        \bottomrule
    \end{tabular}
    \caption{Training Hyperparameters \\ AdamW from \small{\citet{adamw}}}
    \label{tab:traininghyper}
\end{table}

We refer to this pretrained model as \textsc{uspMLM}. For each experiment, \textsc{uspMLM} was fine-tuned on a token classification task. Because the words in the Uspanteko data are already segmented into morphemes, we are able to model this as a token classification task, predicting a gloss for each morpheme. If segmentation were not available, we would have to model the problem with a sequence-to-sequence approach or use some strategy to predict morpheme segmentation. Still, in the token classification approach, the surrounding context for each morpheme is important to making high-quality predictions, and we cannot predict a gloss for each morpheme in a vacuum.

\subsection{Evaluation}
Models are evaluated on both the in-distribution and out-of-distribution evaluation sets. We follow the evaluation strategy used in the SIGMORPHON shared task,
calculating the overall accuracy for every morpheme, ignoring word separators, and requiring glosses to be correctly aligned to morphemes.

\section{Experiments}
For a baseline model, we fine-tune \textsc{uspMLM} on the token classification task. Fine-tuning uses the same hyperparameters listed in \autoref{tab:traininghyper}. We also compare against a naïve strategy where we always select the most common gloss for a morpheme (based on the training data), as well as a strategy that selects a random gloss from the observed glosses for a morpheme in the training data.

We compare our baselines in \autoref{tab:baseline_results}.

\begin{table}[h]
    \centering
    \def\arraystretch{1.5}
    \begin{tabular}{|c|c c|}
    \hline
        \small Model & \small Acc. (Eval ID) & \small Acc. (Eval OOD) \\
        \hline
        Random & 44.4 & 40.6 \\
        Most frequent & \textbf{85.0} & 74.2 \\
        Neural & 84.5 & \textbf{74.6} \\
        \hline
    \end{tabular}
    \caption{Evaluation accuracy on in-distribution and out-of-distribution eval sets for baseline models}
    \label{tab:baseline_results}
\end{table}

All strategies perform worse on the out-of-distribution data. The goal of the following experiments is to improve generalization of the model and thereby close the gap in performance for the ID and OOD evaluation sets. Though the neural model and the naïve model using the most frequent gloss perform similarly, we will conduct experiments with the neural model, which can be more readily manipulated to improve generalization.

\subsection{Optimizing Weight Decay}
Weight decay is important to avoiding overfitting and improving generalization \citep{Loshchilov2017DecoupledWD}, helping reduce variance without sacrificing the representation power of larger models. We fine-tune \textsc{uspMLM} using six different values for weight decay; the results are listed in \autoref{tab:weightdecay_results}.

\begin{table}[h]
    \centering
    \def\arraystretch{1.5}
    \begin{tabular}{|c|c c|}
    \hline
        \small Weight Decay & \small Acc. (Eval ID) & \small Acc. (Eval OOD) \\
        \hline
        0 (Baseline) & 84.5 & 74.6 \\
        0.01 & 84.2 & 73.7 \\
        0.1 & 84.3 & 74.0 \\
        0.5 & \textbf{84.6} & 74.8  \\
        0.75 & \textbf{84.6}  & \textbf{75.1}  \\
        1 & 84.5 & 74.4  \\
        \hline
    \end{tabular}
    \caption{Evaluation accuracy for various weight decay values}
    \label{tab:weightdecay_results}
\end{table}

We find that modifying the weight decay does not significantly affect the accuracy on ID data. However, for OOD data, the best-performing weight decay value of 0.75 achieves a 0.5 percentage point improvement over the baseline. 

Generally, a weight decay of 0 or 0.01 is recommended, so it is interesting that a much larger value of 0.75 is successful in this case. These results could indicate that a more aggressive weight decay allows the model to better generalize to unseen documents, by reducing unnecessary weights and avoiding overfitting. However, the improvement is very small, and its possible that other techniques such as drop out are equally important for mitigating overfitting.

This result is likely heavily dependent on the model architecture, and the optimal weight decay value will vary from model to model. However, increasing weight decay beyond the typical recommendations seems to be an effective strategy.

\subsection{Denoiser}
\subsubsection{Motivation}
Generally, texts from out-of-distribution genres and registers will have more out-of-vocabulary (OOV) tokens in the input. This is the case in our data: the ID eval set has 4.3\% unknown tokens and the OOD eval set has 9.6\% unknown tokens.

Using the best-performing model from the previous section (weight decay of 0.75), we observe that a large portion of the error on the OOD eval set is a result of OOV morphemes in the input. The results of this analysis appear in \autoref{tab:oov_error}.

\begin{table}[h]
    \centering
    \def\arraystretch{1.5}
    \begin{tabular}{|c | c c|}
    \hline
    & Eval (ID) & Eval (OOD) \\
    \hline
    \# OOV Tokens & 527 & 1322 \\
    \# OOV Tokens Incor. & 376 & 854 \\
    Total Incor. & 1910 & 3447 \\
    Total Tokens & 12388 & 13818 \\
    \hline
    \small \# OOV Incor. / Total Incor. & 19.7\% & 24.8\% \\
    \small \# OOV Incor. / Total Tokens & 3.0\% & 6.2\% \\
        \hline
    \end{tabular}
    \caption{Analysis of the error due to out-of-vocabulary (OOV) tokens in the in-distribution (ID) and out-of-distribution (OOD) eval sets}
    \label{tab:oov_error}
\end{table}

OOV tokens contributed 6.2 percentage points to the total error for the OOD data, and only 3.0 points for the ID data. Currently, the best model produces 15.4\% error on the ID data and 24.9\% error on the OOD, with a discrepancy of 9.5 percentage points. Thus, we observe that by reducing the error on OOV tokens, we can decrease a portion of this discrepancy.

\subsubsection{Method}

\begin{figure}[!h]
\centering
\resizebox{0.3\textwidth}{!}{%
\begin{circuitikz}
\tikzstyle{every node}=[font=\LARGE]
\node [font=\LARGE] at (10,5.7) {wi [SEP] qa seboya [SEP] q iik};
\node [font=\LARGE] at (10,8) {wi [SEP] qa [UNK] [SEP] q iik};
\draw [ -Stealth] (10,6.5) -- (10,7.25);
\draw [ -Stealth] (10,9) -- (10,9.75);
\draw [ -Stealth] (10,11.5) -- (10,12.25);
\node [font=\LARGE] at (10,12.75) {EXS [SEP] E1P NOM [SEP] E1P S};
\draw [ color={rgb,255:red,255; green,38; blue,0}, short] (9.75,13.3) .. controls (10.5,12.8) and (10.5,12.8) .. (11,12.4);
\draw [ color={rgb,255:red,255; green,38; blue,0}, short] (9.75,12.4) .. controls (10.5,12.9) and (10.5,12.9) .. (11,13.3);
\draw  (5,16.25) rectangle (15,15);
\draw  (5,11.25) rectangle (15,10);
\draw [ -Stealth] (10,14) -- (10,14.75);
\node [font=\normalsize] at (10,10.5) {Token Classifier};
\node [font=\normalsize] at (10,15.5) {Denoiser};
\node [font=\LARGE] at (10.5,17.75) {S};
\draw [ -Stealth] (10.5,16.5) -- (10.5,17.25);
\node [font=\LARGE] at (10,19.5) {EXS [SEP] E1P S [SEP] E1P S};
\draw [ -Stealth] (10.5,18.5) -- (10.5,19);
\draw [ -Stealth] (7.25,13.75) -- (7.5,18.5);
\draw [ -Stealth] (13,13.75) -- (12.5,18.5);
\end{circuitikz}
}%
\caption{The denoising process. The morpheme "seboya" is OOV, and the token classifier makes an incorrect prediction. However, the denoiser uses observed label sequences to recover the correct gloss, which is substituted into the final prediction.}
\label{fig:denoiser}
\end{figure}
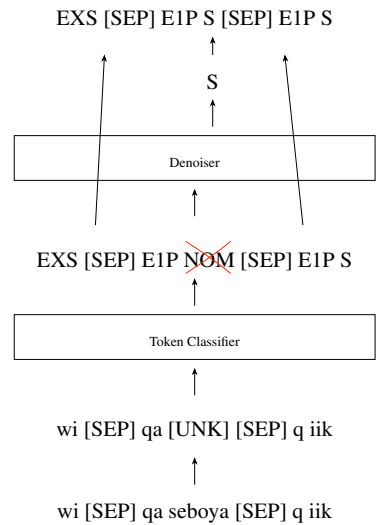

In many languages, morphological patterns are highly regular and structured, and some classes of morphemes (such as agreement morphology) may co-occur in fairly regular ways. We explore the potential of exploiting this fact to make better predictions on unknown morphemes using the other, known morphemes in the sentence. We train a \textbf{denoiser} language model on the gloss sequences in the training set. Then, we use this language model to predict gloss labels for OOV tokens, using the predicted glosses from the token classification model as the input to the denoiser (\autoref{fig:denoiser}).

The denoiser model, \textsc{uspDenoise}, uses the same MLM architecture and training strategy as \textsc{uspMLM}. The model is trained with the hyperparameters in \autoref{tab:traininghyper}, except using a weight decay of 0.01 and 100 epochs.

For inference, we select the examples containing unknown morpheme tokens, and run \textsc{uspDenoise} on the output of the fine-tuned token classification model. Then, we replace the prediction for each OOV morpheme with the prediction from the denoiser. We also experiment with masking the target tokens with the \textsc{MASK} token. We compare with the best-performing model from the previous section in \autoref{tab:denoiser_results}.

\begin{table}[h]
    \centering
    \def\arraystretch{1.5}
    \begin{tabular}{|c|c c|}
    \hline
         Model & \small Acc. (Eval ID) & \small Acc. (Eval OOD) \\
        \hline
        \small No denoiser & 84.6 & 75.1 \\
        \small \makecell{Denoised \\(masked)} & \textbf{84.7} & 74.9 \\
        \small \makecell{Denoised \\(no mask)} & \textbf{84.7} & \textbf{75.3} \\
        \hline
    \end{tabular}
    \caption{Evaluation accuracy for denoiser strategies}
    \label{tab:denoiser_results}
\end{table}

The model using the denoiser without masking tokens shows the best performance, although the improvement is small. In this case, it evidently is difficult to recover the correct token from the surrounding contexts. However, this strategy could still be effective in cases where there are many OOV morphemes or the language is very regular.

\subsection{Self-Supervision}
\subsubsection{Motivation}
Perhaps the most effective way to improve performance on OOD data would simply be to train on OOD data, but in our example scenario this is not feasible. However, we can employ \textbf{iterative pseudo-labeling}, a form of self-supervised learning, to re-train the model using the labels predicted by a prior model \citep{chapelle2009semi}. Iterative pseudo-labeling has been employed in low-resource speech recognition, where additional labeled data is similarly difficult to obtain \citep{kahn2020self}. 

In the context of generalization, iterative pseudo-labeling can help adapt the model to the particular target distribution by re-training the model on predictions for the out-of-domain data. In this way, we can expose the model to the sort of contexts seen in the OOD data without needing additional labeling; retraining the model can also help when the target distribution uses different labeling conventions than the training set.

\subsubsection{Method}

\citet{silovsky2023crosslingual} uses iterative pseudo-labeling to improve performance for low-resource automated speech recognition (ASR) models. We follow their method, described here, hypothesizing that the improvements will be similar for glossing models.

First, we run predictions for our OOD eval set using the best-performing model from the previous section, with a weight decay of 0.75 and denoising. For each sentence, we compute a model confidence value by taking the softmax of the output logits to get the probability value for the most likely gloss at each position and then averaging these probabilities over all glosses in the sequence. We use these confidence values to rank the predictions for every sentence and select some fraction of the predictions with the highest confidence; we experimented with using the top half, third, and quarter of predictions.\footnote{The effectiveness of this approach depends on how well-calibrated the model is.} We pseudo-label these examples with the predicted glosses.

Next, we re-train the trained model using the original training set combined with the selected pseudo-labeled examples. Iterative pseudo-labeling can be run for many iterations if the predictions continue to improve. The results after iterative pseudo-labeling for one iteration, using different fractions of the predictions, are shown in \autoref{tab:pseudolabel_results}.

\begin{table}[h]
    \centering
    \def\arraystretch{1.5}
    \begin{tabular}{|c|c c|}
    \hline
        \small \makecell{Pseudo-labelled\\fraction} & \small Acc. (Eval ID) & \small Acc. (Eval OOD) \\
        \hline
        0 & 84.7 & 75.3 \\
        1/4 & 85.8  & \textbf{76.3} \\
        1/3 & \textbf{85.9}  & 76.2 \\
        1/2 & 85.5 & 75.8 \\
        \hline
    \end{tabular}
    \caption{Evaluation accuracy for models using pseudo-labeling with different fractions of the eval set}
    \label{tab:pseudolabel_results}
\end{table}

We find that the iterative pseudo-labeled models outperform the previous model, with the model using one-quarter of the pseudo-labeled data performing best on the OOD data (with a small tradeoff in ID performance). It seems that selecting a smaller amount of higher-confidence data is more effective than using additional lower-confidence predictions.

Next, we run iterative pseudo-labeling for additional iterations, using the model trained on the top quarter of predictions. In each iteration, we again select the top quarter of predictions, and fine-tune the model. The results after several iterations are given in \autoref{tab:pseudolabel_multiple_results}.

\begin{table}[h]
    \centering
    \def\arraystretch{1.5}
    \begin{tabular}{|c|c c|}
    \hline
        \small \makecell{Iteration} & \small Acc. (Eval ID) & \small Acc. (Eval OOD) \\
        \hline
        0 & 84.7 & 75.3 \\
        1 & 85.8  & 76.3 \\
        2 & \textbf{86.5}  & \textbf{76.9} \\
        3 & 86.3  & 76.8  \\
        \hline
    \end{tabular}
    \caption{Evaluation accuracy after additional iterations of pseudo-labeling}
    \label{tab:pseudolabel_multiple_results}
\end{table}

The second iteration continues to provide performance benefits, but the third iteration shows a small decrease in performance, so we stop iterating and select the model after 2 iterations. While pseudo-labeling initially provides benefits by exposing the model to the target domain, after some iterations the additional noise introduced has detrimental effects. Overall, iterative pseudo-labeling improves the ID accuracy by 1.5 and the OOD accuracy by 1.6 percentage points.

\section{Results}
\autoref{tab:final_results} provides the performance on the held-out, OOD test set using the best model from each step. Each model builds on the previous, so the final model uses all three strategies described in the paper.

\begin{table}[h]
    \centering
    \def\arraystretch{1.5}
    \begin{tabular}{|c|c|}
    \hline
        Model & Acc. (Test OOD) \\
        \hline
        Baseline &  75.5  \\
        WD 0.75 & 76.0 \\
        Denoised & 76.3  \\
        Pseudo-labeled & \textbf{77.5} \\
        \hline
    \end{tabular}
    \caption{Accuracy on held-out test set after applying each technique}
    \label{tab:final_results}
\end{table}

In each step, we use the best trained model from the previous step. We do not iterate pseudo-labeling on the test set, since the test set should have the same distribution as the OOD eval set.

Through weight decay optimization, denoising, and iterative pseudo-labeling, we are able to accomplish an improvement of 2 percentage points in performance on OOD data, with an 8.2\% reduction in overall error.

These techniques also improve performance on the in-distribution eval set, although by a smaller margin than the out-of-distribution eval set. This is desirable, as it narrows the gap between performance on in- and out-of-distribution data, resulting in more predictable model performance.

\section{Discussion}
Although the techniques used in this work do yield performance improvements, generalization in language documentation remains a difficult task, largely due to hard-to-overcome challenges such as unseen morphemes, labels for morphemes that do not appear in the training set, and ambiguity in labeling. 

Weight decay optimization, like all forms of hyperparameter tuning, is highly situation-dependent and requires good evaluation. Generally, avoiding overfitting and minimizing variance is critical to generalization in documentation, where the training sets may represent only a small fraction of the distribution of possible texts.

Denoising is a promising strategy for making high-quality predictions on completely unknown morphemes, using the surrounding context. This approach may be particularly useful in a human-in-the-loop situation, where the denoiser provides several top guesses for an unknown morpheme, and a human annotator can select between the options, allowing for easier annotation and possibly active learning \citep{palmer2009evaluating}. Denoising will likely show more robust performance for languages with highly structured and productive morphological systems and relationships such as agreement and regular word order.

Some aspects of Uspanteko morphology are productive and structured. For example, verbs can take multiple affixes, both prefixes and suffixes, and these occur in a predictable order, according to a morphological pattern. At the same time, the language also has relatively flexible classes of morphemes, allowing non-verbal stems to act as predicates \citep{coon_mayan_2016}, taking some of the same morphology as seen on verb stems. This flexibility may have decreased the utility of the denoising approach, as unseen stems appearing in verbal positions could be verbal or non-verbal morphemes, with no clear distinction.

Iterative pseudo-labeling similarly shows only a small improvement. In these experiments, the OOD texts still share fairly similar contexts and labeling strategies with the training set, as evidenced by the perplexity values. However, in a case where the unseen texts are more dissimilar to the training set, this strategy could be more effective at tuning the model to the particular target distribution.

\section{Future Research}
This work presents a preliminary exploration into generalization for documentation models, and much work remains to be done. Documentation data for even the most widely-spoken languages is limited, yet robust generalization from the training set is crucial for improving usability.

One promising approach for creating more robust documentation models is through cross-lingual transfer that utilizes the morphological similarities between languages. \citet{he-etal-2023-sigmorefun} demonstrates that this approach can effect performance improvements on in-distribution data, and it would likely benefit out-of-distribution data as well.

Another technique for avoiding overfitting and improving generalization is ensuring models focus on linguistic information, relying less on semantic patterns that may lead to spurious generalizations. This could involve morphologically inspired loss functions, data augmentation using rule-based systems, or pretraining on other linguistic tasks.

\section{Conclusion}
In this work, we presented three strategies for improving generalization of interlinear glossed text generation models, which to our knowledge are novel approaches to the problem. We use weight decay optimization, denoising, and iterative pseudo-labeling, finding that iterative pseudo-labeling provides the greatest improvement in performance. Overall, our best model achieves a 2\% improvement from the baseline on a test set representing texts of unseen genres. We also investigate the discrepancy in performance between in- and out-of-distribution data, finding that out-of-vocabulary morphemes and differences in context are key sources of error. We hope these approaches can inspire future work in improving generalization for documentation models, which is difficult but critical to the usability of automated documentation systems in real-world projects.

\section{Limitations}
This research was conducted testing on a single language and corpus, and the effectiveness of each approach may vary with the language used. Additionally, this work focused on glossing morphemes, provided words have already been segmented into morphemes. This is often not the case for IGT data, and segmentation remains a difficult problem.

The experiments utilized a single model architecture for consistency, but other architectures might show different performance. We used a small transformer architecture due to the size of the training dataset; a deeper network might show different results.

We focused on experimenting with texts of unseen genres as our out-of-distribution data, but this is only one form of generalization. Other types of OOD data include data from other speakers or communities, dialects of a language, and data from different documentation projects. 

\section{Ethical Considerations}
When working with projects that affect language communities, we should always strive to avoid a colonialist approach, and we should bear in mind that language data does not exist in a vacuum, but is the product of human experience \citep{bird_decolonising_2020}. Documentation projects should never be undertaken without the consent and cooperation of the relevant language community.

Generalization is desirable in order to produce more valuable documentation systems, but it can also cause the homogenization of language, which can particularly affect speakers of less widely spoken dialects.

Training large transformer models requires a large amount of computation and thus incurs an unavoidable carbon cost \citep{bender_dangers_2021}, and thus we aimed to keep the architectures as small as possible. Minimizing the environmental impact of machine learning is a critical ongoing area of research.

\section{Acknowledgements}
We thank the anonymous reviewers for their useful suggestions and feedback, as well as the LECS Lab at the University of Colorado. This material is based upon work supported by the National Science Foundation under Grant No. 2149404, “CAREER: From One Language to Another”. Any opinions, findings, and conclusions or recommendations expressed in this material are those of the authors and do not necessarily reflect the views of the National Science Foundation.

\bibliography{anthology,custom}
\bibliographystyle{acl_natbib}

\appendix

\section{GenBench Evaluation Card}
\begin{figure}[h]
    \centering
    
\newcommand{\tabularwidth}{\columnwidth}

\newcommand{\expone}{$\square$}
        
\renewcommand{\arraystretch}{1.1}         
\setlength{\tabcolsep}{0mm}         
\begin{tabular}{|p{\tabularwidth}<{\centering}|}         
\hline
               
\rowcolor{gray!60}               
\textbf{Motivation} \\               
\footnotesize
\begin{tabular}{p{0.25\tabularwidth}<{\centering} p{0.25\tabularwidth}<{\centering} p{0.25\tabularwidth}<{\centering} p{0.25\tabularwidth}<{\centering}}                        
\textit{Practical} & \textit{Cognitive} & \textit{Intrinsic} & \textit{Fairness}\\
\expone\hspace{0.8mm}		
& 		
& 		
& 		

\vspace{2mm} \\
\end{tabular}\\
               
\rowcolor{gray!60}               
\textbf{Generalisation type} \\               
\footnotesize
\begin{tabular}{m{0.17\tabularwidth}<{\centering} m{0.20\tabularwidth}<{\centering} m{0.14\tabularwidth}<{\centering} m{0.17\tabularwidth}<{\centering} m{0.18\tabularwidth}<{\centering} m{0.14\tabularwidth}<{\centering}}                   
\textit{Compo- sitional} & \textit{Structural} & \textit{Cross Task} & \textit{Cross Language} & \textit{Cross Domain} & \textit{Robust- ness}\\
& 		
& 		
& 		
& \expone\hspace{0.8mm}		
& \expone\hspace{0.8mm}		

\vspace{2mm} \\
\end{tabular}\\
             
\rowcolor{gray!60}             
\textbf{Shift type} \\             
\footnotesize
\begin{tabular}{p{0.25\tabularwidth}<{\centering} p{0.25\tabularwidth}<{\centering} p{0.25\tabularwidth}<{\centering} p{0.25\tabularwidth}<{\centering}}                        
\textit{Covariate} & \textit{Label} & \textit{Full} & \textit{Assumed}\\  
\expone\hspace{0.8mm}		
& 		
& 		
& 		

\vspace{2mm} \\
\end{tabular}\\
             
\rowcolor{gray!60}             
\textbf{Shift source} \\             
\footnotesize
\begin{tabular}{p{0.25\tabularwidth}<{\centering} p{0.25\tabularwidth}<{\centering} p{0.25\tabularwidth}<{\centering} p{0.25\tabularwidth}<{\centering}}                          
\textit{Naturally occuring} & \textit{Partitioned natural} & \textit{Generated shift} & \textit{Fully generated}\\
\expone\hspace{0.8mm}		
& 		
& 		
& 		

\vspace{2mm} \\
\end{tabular}\\
             
\rowcolor{gray!60}             
\textbf{Shift locus}\\             
\footnotesize
\begin{tabular}{p{0.25\tabularwidth}<{\centering} p{0.25\tabularwidth}<{\centering} p{0.25\tabularwidth}<{\centering} p{0.25\tabularwidth}<{\centering}}                         
\textit{Train--test} & \textit{Finetune train--test} & \textit{Pretrain--train} & \textit{Pretrain--test}\\
\expone\hspace{0.8mm}		
& 		
& 		
& 		

\vspace{2mm} \\
\end{tabular}\\

\hline
\end{tabular}

    \caption{GenBench evaluation card as described in \citet{hupkes2023taxonomy}}
\end{figure}

\end{document}